# Neural-Symbolic Argumentation Mining: An Argument in Favor of Deep Learning and Reasoning

Andrea Galassi[1]*, Kristian Kersting[2], Marco Lippi[3], Xiaoting Shao[2] and Paolo Torroni[1]

[1] Department of Computer Science and Engineering, University of Bologna, Bologna, Italy, [2] Computer Science Department and Centre for Cognitive Science, TU Darmstadt, Darmstadt, Germany, [3] Department of Sciences and Methods for Engineering, University of Modena and Reggio Emilia, Reggio Emilia, Italy

Deep learning is bringing remarkable contributions to the field of argumentation mining, but the existing approaches still need to fill the gap toward performing advanced reasoning tasks. In this position paper, we posit that neural-symbolic and statistical relational learning could play a crucial role in the integration of symbolic and sub-symbolic methods to achieve this goal.

## 1. INTRODUCTION

The goal of argumentation mining (AM) is to automatically extract arguments and their relations from a given document (Lippi and Torroni, 2016). The majority of AM systems follows a pipeline scheme, starting with simpler tasks, such as argument component detection, down to more complex tasks, such as argumentation structure prediction. Recent years have seen the development of a large number of techniques in this area, on the wake of the advancements produced by deep learning on the whole research field of natural language processing (NLP). Yet, it is widely recognized that the existing AM systems still have a large margin of improvement, as good results have been obtained with some genres where prior knowledge on the structure of the text eases some AM tasks, but other genres, such as legal cases and social media documents still require more work (Cabrio and Villata, 2018). Performing and understanding argumentation requires advanced reasoning capabilities, which are natural human skills, but are difficult to learn for a machine. Understanding whether a given piece of evidence supports a given claim, or whether two claims attack each other, are complex problems that humans can address thanks to their ability to exploit commonsense knowledge, and to perform reasoning and inference. Despite the remarkable impact of deep neural networks in NLP, we argue that these techniques alone will not suffice to address such complex issues.

We envisage that a significant advancement in AM could come from the combination of symbolic and sub-symbolic approaches, such as those developed in the Neural Symbolic (NeSy) (Garcez et al., 2015) or Statistical Relational Learning (SRL) (Getoor and Taskar, 2007; De Raedt et al., 2016; Kordjamshidi et al., 2018) communities. This issue is also widely recognized as one of the major challenges for the whole field of artificial intelligence in the coming years (LeCun et al., 2015).

In computational argumentation, structured arguments have been studied and formalized for decades using models that can be expressed in a logic framework (Bench-Capon and Dunne, 2007). At the same time, AM has rapidly evolved by exploiting state-of-the-art neural architectures coming from deep learning. So far, these two worlds have progressed largely independently of each other.

Only recently, a few works have taken some steps toward the integration of such methods, by applying techniques combining sub-symbolic classifiers with knowledge expressed in the form of rules and constraints to AM. For instance, Niculae et al. (2017) adopted structured support vector machines and recurrent neural networks to collectively classify argument components and their relations in short documents, by hard-coding contextual dependencies and constraints of the argument model in a factor graph. A joint inference approach for argument component classification and relation identification was instead proposed by Persing and Ng (2016), following a pipeline scheme where integer linear programming is used to enforce mathematical constraints on the outcomes of a first-stage set of classifiers. More recently, Cocarascu and Toni (2018) combined a deep network for relation extraction with an argumentative reasoning system that computes the dialectical strength of arguments, for the task of determining whether a review is truthful or deceptive.

We propose to exploit the potential of both symbolic and sub-symbolic approaches for AM, by combining both results in systems that are capable of modeling knowledge and constraints with a logic formalism, while maintaining the computational power of deep networks. Differently from existing approaches, we advocate the use of a logic-based language for the definition of contextual dependencies and constraints, independently of the structure of the underlying classifiers. Most importantly, the approaches we outline do not exploit a pipeline scheme, but rather perform joint detection of argument components and relations through a single learning process.

## 2. MODELING ARGUMENTATION WITH PROBABILISTIC LOGIC

Computational argumentation is concerned with modeling and analyzing argumentation in the computational settings of artificial intelligence (Bench-Capon and Dunne, 2007; Simari and Rahwan, 2009). The formalization of arguments is usually addressed at two levels. At the argument level, the definition of formal languages for representing knowledge and specifying how arguments and counterarguments can be constructed from that knowledge is the domain of *structured argumentation* (Besnard et al., 2014). In structured argumentation, the premises and claim of the argument are made explicit, and their relationships are formally defined. However, when the discourse consists of multiple arguments, such arguments may conflict with one another and result in logical inconsistencies. A typical way of dealing with such inconsistencies is to identify sets of arguments that are mutually consistent and that collectively defeat their "attackers." One way to do that is to abstract away from the internal structure of arguments and focus on the higher-level relations among arguments: a conceptual framework known as *abstract argumentation* (Dung, 1995).

Similarly to structured argumentation, AM too builds on the definition of an argument model, and aims to identify parts of the input text that can be interpreted as argument components (Lippi and Torroni, 2016). For example, if we take a basic claim/evidence argument model, possible tasks could be claim detection (Aharoni et al., 2014; Lippi and Torroni, 2015), evidence detection (Rinott et al., 2015), and the prediction of links between claim and evidence (Niculae et al., 2017; Galassi et al., 2018). However, in structured argumentation the formalization of the model is the basis for an inferential process, whereby conclusions can be obtained starting from premises. In AM, instead, an argument model is usually defined in order to identify the target classes, and in some isolated cases to express relations, for instance among argument components (Niculae et al., 2017), but not for producing inferences that could help the AM tasks.

The languages of structured argumentation are logic-based. An influential structured argumentation system is *deductive argumentation* (Besnard and Hunter, 2001), where premises are logical formulae, which entail a claim, and entailment may be specified from a range of base logics, such as classical logic or modal logic. In *assumption-based argumentation* (Dung et al., 2009) instead arguments correspond to assumptions, which like in deductive systems prove a claim, and attacks are obtained via a notion of contrary assumptions. Another powerful framework is *defeasible logic programming* (DeLP) (García and Simari, 2004), where claims can be supported using strict or defeasible rules, and an argument supporting a claim is warranted if it defeats all its counterarguments. For example, that a cephalopod is a mollusc could be expressed by a strict rule, such as:

$$mollusc(X) \leftarrow cephalopod(X)$$

as these notions belong to an artificially defined, incontrovertible taxonomy. However, since in nature not all molluscs have a shell, and actually cephalopods are molluscs without a shell, rules used to conclude that a given specimen has or does not have a shell are best defined as defeasible. For instance, one could say:

$$has\_shell(X) \prec mollusc(X)$$
$$\sim has\_shell(X) \prec cephalopod(X)$$

where $\prec$ denotes defeasible inference.

The choice of logic notwithstanding, rules offer a convenient way to describe argumentative inference. Moreover, depending on the application domain, the document genre, and the employed argument model, different constraints and rules can be enforced on the structure of the underlying network of arguments. For example, if we adopt a DeLP-like approach, strict rules can be used to define the relations among argument components, and defeasible rules to define context knowledge. For example, in a hypothetical claim-premise model, support relations may be defined exclusively between a premise and a claim. Such structural properties could be expressed by the following strict rules:

$$claim(Y) \leftarrow supports(X, Y)$$
$$premise(X) \leftarrow supports(X, Y)$$

whereby if $X$ supports $Y$, then $X$ is a claim and $Y$ is a premise. As another abstract example, two claims based on the same premise may not attack each other:

$$\sim attacks(Y1, Y2) \leftarrow supports(X, Y1) \wedge supports(X, Y2)$$

As an example of defeasible rules, consider instead the context information about a political debate, where a republican candidate, R, faces a democrat candidate, D. Then one may want to use the knowledge that R's claims and D's claims are likely to attack each other:

$$attacks(Y1, Y2) \prec auth(Y1, R) \wedge rep(R) \wedge auth(Y2, D) \wedge dem(D)$$

where predicate $auth(A, B)$ denotes that claim A was made by B. There exist many non-monotonic reasoning systems that integrate defeasible and strict inference. However, an alternative approach that may successfully reconcile the computational argumentation view and the AM view is offered by probabilistic logic programming (PLP). PLP combines the capability of logic to represent complex relations among entities with the capability of probability to model uncertainty over attributes and relations (Riguzzi, 2018). In a PLP framework, such as PRISM (Sato and Kameya, 1997), LPAD (Vennekens et al., 2004), or ProbLog (Raedt et al., 2007), defeasible rules may be expressed by rules with a probability label. For instance, in LPAD syntax, one could write:

$$attacks(Y1, Y2) : 0.8 \leftarrow auth(Y1, R) \wedge rep(R) \wedge auth(Y2, D)$$
$$\wedge dem(D)$$

to express that the above rule holds in 80% of cases. In this example, 0.8 could be interpreted as a weight or score suggesting how likely the given inference rule is to hold. In more recent approaches, such weights could be learned directly from a collection of examples, for example by exploiting likelihood maximization in a *learning from interpretations* setting (Gutmann et al., 2011) or by using a generalization of expectation maximization applied to logic programs (Bellodi and Riguzzi, 2013).

From a higher-level perspective, rules could be exploited also to model more complex relations between arguments or even to encode argument schemes (Walton et al., 2008), for example to assess whether an argument is defeated by another, on the basis of the strength of its premises and claims. This is an even more sophisticated reasoning task, which yet could be addressed within the same kind of framework described so far.

## 3. COMBINING SYMBOLIC AND SUB-SYMBOLIC APPROACHES

The usefulness of deep networks has been tested and proven in many NLP tasks, such as machine translation (Young et al., 2018), sentiment analysis (Zhang et al., 2018a), text classification (Conneau et al., 2017; Zhang et al., 2018b), relations extraction (Huang and Wang, 2017), as well as in AM (Cocarascu and Toni, 2017, 2018; Daxenberger et al., 2017; Galassi et al., 2018; Lauscher et al., 2018; Lugini and Litman, 2018; Schulz et al., 2018). While a straightforward approach to exploit domain knowledge in AM is to apply a set of hand-crafted rules on the output of some first stage classifier (such as a neural network), NeSy or SRL approaches can directly enforce (hard or soft) constraints *during training*, so that a solution that does not satisfy them is penalized, or even ruled out. Therefore, if a neural network is trained to classify argument components, and another one[1] is trained to detect links between them, additional global constraints can be enforced to adjust the weights of the networks toward admissible solutions, as the learning process advances. Systems like DeepProbLog (Manhaeve et al., 2018), Logic Tensor Networks (Serafini and Garcez, 2016), or Grounding-Specific Markov Logic Networks (GS-MLN) (Lippi and Frasconi, 2009), to mention a few, enable such a scheme.

By way of illustration, we report how to implement one of the cases mentioned in section 2 with DeepProbLog and with GS-MLNs. By extending the ProbLog framework, DeepProbLog allows to introduce the following kind of construct:

$$nn(m, \vec{t}, \vec{u}) :: q(\vec{t}, u_1); \ldots; q(\vec{t}, u_n).$$

The effect of the construct is the creation of a set of ground probabilistic facts, whose probability is assigned by a neural network. This mechanism allows us to delegate to a neural network $m$ the classification of a set of predicates $q$ defined by some input features $\vec{t}$. The possible classes are given by $\vec{u}$. Therefore, in the AM scenario, it would be possible, for example, to exploit two networks $m\_t$ and $m\_r$ to classify, respectively, the type of a potential argumentative component and the potential relation between two components. **Figure 1A** shows the corresponding DeepProbLog code. These predicates could be easily integrated within a probabilistic logic program designed for argumentation, so as to model (possibly weighted) constraints, rules, and preferences, such as those described in section 2. **Figure 1B** illustrates one such possibility.

The same scenario can be modeled using GS-MLNs. In the Markov logic framework, first-order logic rules are associated with a real-valued weight. The higher the weight, the higher the probability that the clause is satisfied, other things being equal. The weight could possibly be infinite, to model hard constraints. In the GS-MLN extension, different weights can be associated to different groundings of the same formula, and such weights can be computed by neural networks. Joint training and inference can be performed, as a straightforward extension of the classic Markov logic framework (Lippi and Frasconi, 2009). **Figure 2** shows an example of a GS-MLN used to model the AM scenario that we consider. Here, the first three rules model grounding-specific clauses (the dollar symbol indicating a reference to a generic vector of features describing the example) whose weights depend on the specific groundings (variables x and y); the three subsequent rules are hard constraints (ending with a period); the final rule is a classic Markov logic rule, with a weight attached to the first-order clause.

---

[1] In a multi-task setting, instead of two networks, the same network could be used to perform both component classification and link detection at the same time. In AM, this approach has already been successfully explored (Eger et al., 2017; Galassi et al., 2018).

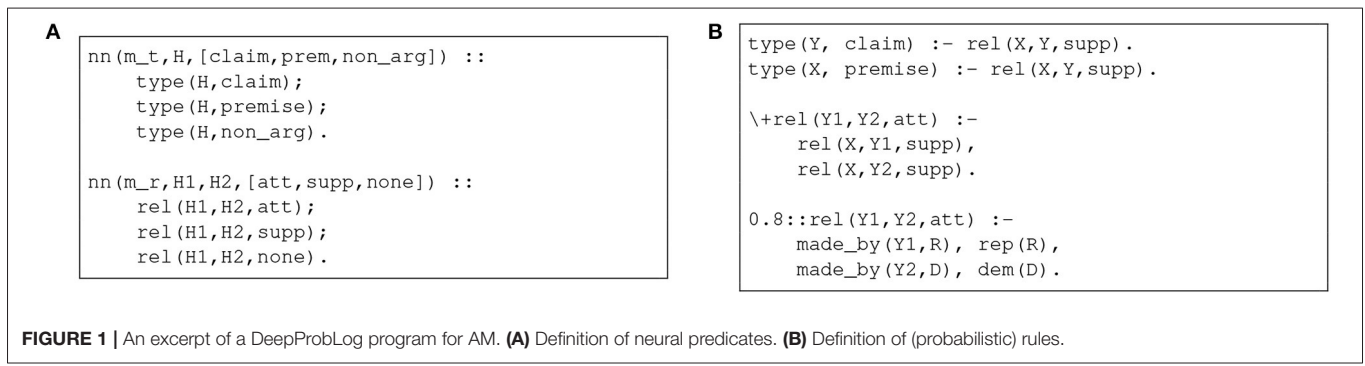

FIGURE 1 | An excerpt of a DeepProbLog program for AM. **(A)** Definition of neural predicates. **(B)** Definition of (probabilistic) rules.

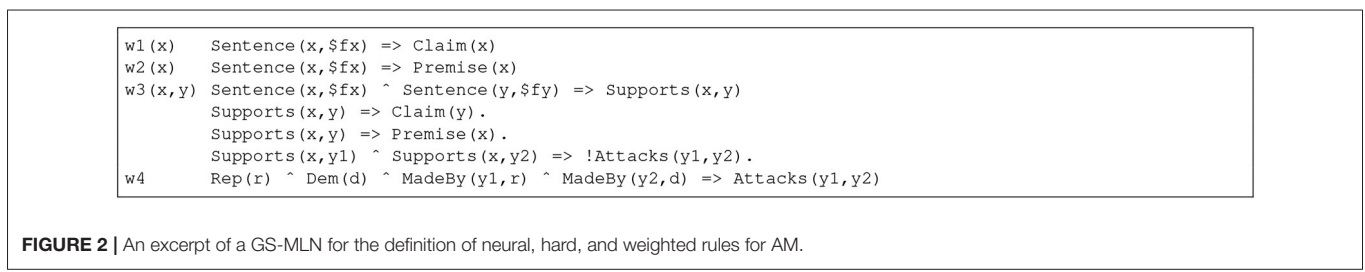

FIGURE 2 | An excerpt of a GS-MLN for the definition of neural, hard, and weighted rules for AM.

The kind of approach hereby described strongly differs from the existing approaches in AM. Whereas, Persing and Ng (2016) exploit a pipeline scheme to apply the constraints to the predictions made by deep networks at a first stage of computation, the framework we propose can perform a joint training, which includes the constraints within the learning phase. This can be viewed as an instance of *Constraint Driven Learning* (Chang et al., 2012) and its continuous counterpart, posterior regularization (Ganchev et al., 2010), where multiple signals contribute to a global decision, by being pushed to satisfy expectations on the global decision. Differently from the work by Niculae et al. (2017), who use factor graphs to encode inter-dependencies between random variables, our approach enables to exploit the interpretable formalism of logic to represent rules. Moreover, the models of NeSy and SRL are typically able to learn the weights or the probabilities of the rules, or even to learn the rules themselves, thus addressing a *structure learning* task.

## 4. DISCUSSION

After many years of growing interest and remarkable results, time is ripe for AM to move forward in its ability to support complex arguments. To this end, we argue that research in this area should aim at combining sub-symbolic and symbolic approaches, and that several state-of-the-art ML frameworks already provide the necessary ground for such a leap forward.

The combination of such approaches will leverage different forms of abstractions that we consider essential for AM. On the one hand, (probabilistic) logical representations enable to specify AM systems in terms of data, world knowledge and other constraints, and to express uncertainties at a logical and conceptual level rather than at the level of individual random variables. This would make AM systems easier to interpret—a feature that is now becoming a need for AI in general (Guidotti et al., 2018)—since they could help explaining the logic and the reasons that lead them to produce their arguments, while still dealing with the uncertainties stemming from the data and the (incomplete) background knowledge. On the other hand, AM is too complex to fully specify the distributions of random variables and their global (in)dependency structure a priori. Sub-symbolic models can harness such a complexity by finding the right, general outline, in the form of computational graphs, and processing data.

In order to fully exploit the potential of this joint approach, clearly many challenges have to be faced. First of all, several languages and frameworks for NeSy and SRL exist, each with its own characteristics in terms of both expressive power and efficiency. In this sense, AM would represent an ideal test-bed for such frameworks, by presenting a challenging, large-scale application domain where the exploitation of background knowledge could play a crucial role to boost performance. Inference in this kind of models is clearly an issue, thus AM would provide additional benchmarks for the development of efficient algorithms, both in terms of memory consumption and running time. Finally, although there are already several NeSy and SRL frameworks available, being these research areas still relatively young and in rapid development, their tools are not yet mainstream. Here, an effort is needed in integrating such tools with state-of-the-art neural architectures for NLP.

## AUTHOR CONTRIBUTIONS

All authors listed have made a substantial, direct and intellectual contribution to the work, and approved it for publication.

## FUNDING

This work has been partially supported by the H2020 Project AI4EU (G.A. 825619), by the DFG project CAML (KE 1686/3-1, SPP 1999), and by the RMU project DeCoDeML.

## ACKNOWLEDGMENTS

We thank the reviewers for their comments and contributions, which have increased the quality of this work. A preliminary version of this manuscript has been released as a Pre-Print at Galassi et al. (2019).